\pdfoutput=1

\documentclass[11pt]{article}

\usepackage[final]{acl}

\usepackage{amsmath}
\usepackage{amssymb}
\usepackage{times}
\usepackage{latexsym}
\usepackage{kotex}
\usepackage{booktabs}
\usepackage{multirow}
\usepackage{lscape}
\usepackage{array}
\usepackage{longtable}
\usepackage{color}
\usepackage{tabularray}
\usepackage{arydshln}
\usepackage{subcaption}
\usepackage{tabularx}

\usepackage{tikz}
\usetikzlibrary{arrows.meta}  

\usepackage[T1]{fontenc}

\usepackage[utf8]{inputenc}

\usepackage{microtype}

\usepackage{inconsolata}

\usepackage{graphicx}

\title{Language Chain in Alignment: \\ Cross-lingual Ranking Preference Optimization}

\author{
    Seungyoon Lee, Minhyuk Kim, Jungseob Lee, Heuiseok Lim\thanks{Corresponding author} \\ 
    Department of Computer Science and Engineering, Korea University \\
    \texttt{\{dltmddbs100, mhkim0929, omanma1928, limhseok\}@korea.ac.kr}
}

\begin{document}
\maketitle
\begin{abstract}
The alignment of Large Language Models heavily relies on English-centric high-quality preference data, which often leads to suboptimal performance in other languages. In this paper, we propose Cross-lingual Ranking Preference Optimization~(CRPO), a novel framework that leverages robust preference knowledge from English to facilitate preference alignment in the target language. We design a hierarchical structure within parallel preference pairs across the target language and English to jointly optimize intra- and inter-lingual preferences, thereby enhancing language adaptation and output quality. Building on the LambdaLoss framework, CRPO goes beyond the binary comparison based optimization by providing a relative ranking signal across multiple candidate responses. Our experiments across five languages with varying resource scales demonstrate that CRPO consistently outperforms standard approaches in both instruction-following and knowledge utilization capability. Notably, the robust performance gains observed across various weighting schemes further validate the empirical effectiveness of our hierarchical design in a multilingual setup. Furthermore, our findings highlight that CRPO significantly improves both reward margins and the log-probability of desirable responses, contributing to a more stable preference manifold for cross-lingual alignment. Our code is available at \url{https://github.com/dltmddbs100/CRPO}.
\end{abstract}

\section{Introduction}
Integrating human feedback through the alignment process has emerged as a critical paradigm for optimizing the instruction-following capabilities of Large Language Models~(LLMs) and grounding their objective functions in human value systems~\cite{stiennon2020learning,askell2021general,hong2024orpo,meng2024simpo}. In particular, Direct Preference Optimization~(DPO)~\cite{rafailov2023direct} has streamlined the complex Reinforcement Learning from Human Feedback~(RLHF)~\cite{ouyang2022training} pipeline, effectively guiding models toward preferred responses while improving helpfulness and utility~\cite{kirk2023understanding,pang2024iterative,tu2025enhancing}. However, these alignment gains remain largely confined to English-centric contexts, leaving consistency and response accuracy in other languages as a formidable challenge. This disparity often leads to undesirable behavior, including input-output language inconsistency and degraded performance for other language queries~\cite{lai-etal-2023-chatgpt,puttaparthi2023comprehensive,huang-etal-2023-languages,wu2024not,marchisio-etal-2024-understanding}.

\begin{figure}[t]
\centering 
\includegraphics[width=\linewidth]{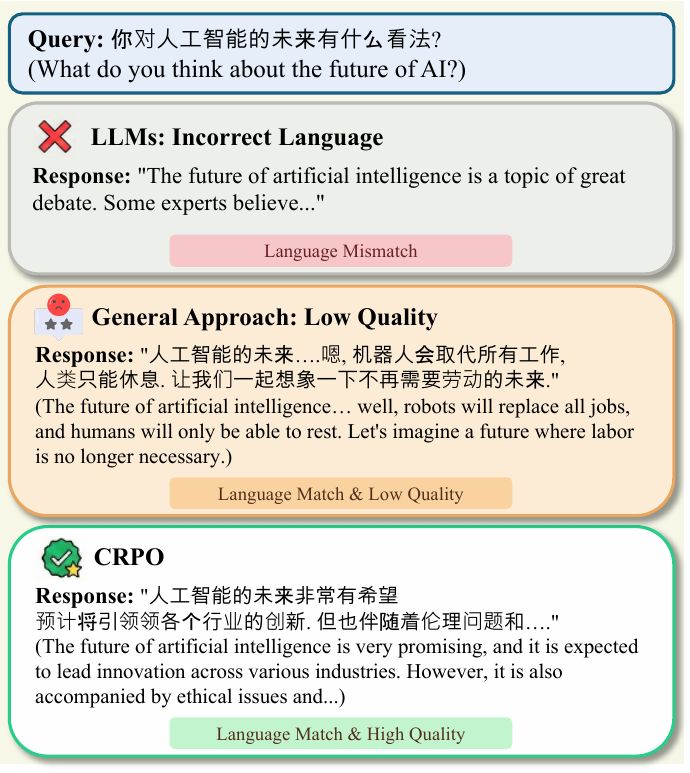}
\caption{Example of model responses to multilingual prompt after alignment tuning.}
\label{fig:intro}
\end{figure}

Existing approaches to multilingual alignment typically rely on expensive
human annotation or on integrating target language data into English-dominant corpora~\cite{lai2023okapi,zhao2024llama,ranaldi-etal-2024-empowering,lai-etal-2024-llms}. However, these approaches are limited by their inability to leverage the model's inherent English capabilities for cross-lingual alignment. By leaving this internalized English preference knowledge decoupled from the target language, they struggle to achieve effective cross-lingual alignment.

In this paper, we propose \textbf{C}ross-lingual \textbf{R}anking \textbf{P}reference \textbf{O}ptimization~(CRPO), a novel alignment framework that aligns the preference distribution of a target language by leveraging the model's inherent preference knowledge in English. CRPO moves beyond conventional binary comparisons within a single language; instead, we adopt an alignment strategy based on a hierarchical ranking structure that integrates two core dimensions: language consistency and response quality. As illustrated in Figure~\ref{fig:intro}, while LLMs often suffer from language confusion~(e.g., responding in English to a non-English prompt) and existing approaches generate low-quality outputs, CRPO guides the model toward generating high-quality responses appropriate for the target language query.

To achieve this, we transplant the ``Learning-to-Rank''~(LTR)~\cite{liu2009learning} philosophy, which optimizes relative importance among multiple candidates, into the cross-lingual alignment training. To be specific, we design a hierarchical signal based on parallel preference pairs between English and the target language. This design facilitates alignment of target language preferences by referring to well-defined preference distributions, while inducing the generation of target language responses rather than English.

Our experiments show that CRPO achieves substantial improvements in instruction-following and knowledge-intensive tasks. Furthermore, our comparative analysis of reward and log-likelihood distributions demonstrates that CRPO leads to a more accurate likelihood preference of winning and losing responses on a held-out test set, which in turn translates to a better policy model for the target language. Our contributions are as follows:
\begin{itemize}
  \item We propose CRPO, a novel framework that optimizes the hierarchical priority between language and quality by integrating ranking principles into cross-lingual preference alignment.
  \item Through an analysis of reward and probability distributions, we find that CRPO more effectively steers the preference distribution of the target language toward a desirable alignment compared to existing methodologies.
  \item We demonstrate that our cross-lingual preference hierarchy structure itself, which leverages the relationship between English and the target language, induces significant performance gains in target language alignment.
\end{itemize}

\section{Related Work}
Enhancing the multilingual capabilities of LLMs is a critical endeavor to ensure that global users can benefit from technological advancements without language barriers~\cite{costa2022no,zhao2024llama,huang2024survey,li2024improving}. Most research primarily focused on machine translating English datasets or training on heterogeneous mixtures of multilingual corpora~\cite{workshop2022bloom,muennighoff-etal-2023-crosslingual,lai2023okapi,gao-etal-2024-multilingual,kew-etal-2024-turning}. However, these approaches often require large-scale datasets or suffer from persistent language bias~\cite{aggarwal-etal-2024-maple,li2024quantifying,lee-etal-2025-cross}.

Recent studies seeking to mitigate these limitations have increasingly explored extending alignment algorithms proven in English to multilingual contexts~\cite{yang2024language,dang-etal-2024-rlhf}. In prior work, \citet{wu2024reuse} facilitates zero-shot cross-lingual transfer in reward model training, aiming to build robust multilingual reward systems. MAPO~\cite{she-etal-2024-mapo} targets improvement in the multilingual reasoning domain using reward signals derived from external translation models. Similarly, MPO~\cite{zhao-etal-2025-mpo} directly optimizes the reward margin gap across languages to enhance safety. While effective, these approaches rely on external dependencies, such as online machine translation pipelines, and are limited to specific domains. Another line of work operates directly on the policy model using implicit binary comparisons~\cite{yang-etal-2025-implicit,lee-etal-2025-cross}, yet they remain constrained by binary optimization, failing to capture complex preference distributions.

Although some works have explored list-based ranking to move beyond binary comparisons, these studies primarily focus on improving quality or safety and are strictly limited to English-centric contexts~\cite{yuan2023rrhf,song2024preference,liu-etal-2025-lipo}. In contrast, CRPO distinguishes itself by optimizing a cross‑lingual hierarchical ranking structure that jointly models language and quality. By enabling simultaneous modeling of linguistic and quality hierarchies, CRPO leads to meaningful improvements in target language capabilities.

\section{CRPO Framework}
Our approach is established on the DPO framework, extending its pairwise alignment capabilities into a ranking paradigm specifically designed for cross-lingual contexts.

\subsection{Preliminary: Direct Preference Optimization}
\label{sec:dpo}
DPO~\cite{rafailov2023direct} is a stable and efficient method to align LLMs with human preferences without the need for training an explicit reward model or using complex reinforcement learning pipelines such as PPO~\cite{schulman2017proximal}:
\begin{equation}
\begin{aligned}
&\mathcal{L}_{\text{DPO}}(\pi_\theta; \pi_{\text{ref}})= - \mathbb{E}_{(x,y_w,y_l) \sim \mathcal{D}} \\
&\left[ \log \sigma \left( \beta \log \frac{\pi_\theta(y_w | x)}{\pi_{\text{ref}}(y_w | x)} - \beta \log \frac{\pi_\theta(y_l | x)}{\pi_{\text{ref}}(y_l | x)} \right) \right]
\label{eq:dpo}
\end{aligned}
\end{equation}
where $\pi_{\theta}$ is the policy model, $\pi_{\text{ref}}$ is the reference policy given $(x,y_{w},y_{l})$ are preference pairs which are the prompt, chosen response, and rejected response from the preference dataset. By minimizing the given loss, the model increases the margin between the preferred response $y_{w}$ relative to $y_{l}$.

\subsection{LambdaLoss Ranking Objective}
\label{sec:lambdaloss}
Given that our aim is to encourage cross-lingual adaptation alongside robust preference alignment, the binary formulation of DPO is inherently limited, as it cannot accommodate multiple optimization spheres. To address this, we draw inspiration from LambdaLoss~\cite{wang2018lambdaloss}, a type of Learning to Rank~(LTR)~\cite{liu2009learning} framework that enables joint optimization of multiple items across a global ranking structure.

This approach allows the model to optimize well-founded ranking metrics, such as Discounted Cumulative Gain~(DCG)~\cite{burges2006learning}, by weighting pairwise transitions based on their impact on the overall list. Specifically, for a prompt $x$ and a set of candidate responses $\mathbf{y} = \{y_1, \dots, y_n\}$, the training objective of our ranking scheme is defined as:
\begin{equation}
\begin{aligned}
&\mathcal{L}_{\text{lambda}}(\pi_{\theta}; \pi_{\text{ref}}, \beta) = \\
&\mathbb{E}_{(x, \mathbf{y}, \psi) \sim \mathcal{D}} \left[ \sum_{\psi_i > \psi_j} \Delta_{i,j} \log(1 + e^{-(s_i - s_j)}) \right]
\label{eq:lambda_loss}
\end{aligned}
\end{equation}
where $s_i = \beta \log \frac{\pi_\theta(y_i | x)}{\pi_{\text{ref}}(y_i | x)}$ represents the implicit reward score of response $y_i$. The term $\Delta_{i,j}$ is the Lambda weight, which determines the importance of the pair $(y_i, y_j)$ based on their positions in the ranking:
\begin{equation}
\Delta_{i,j} = |G_i - G_j| \cdot \left| \frac{1}{D(\tau(i))} - \frac{1}{D(\tau(j))} \right|
\label{eq:lambda_weight}
\end{equation}
In this formulation, $G$ is the gain function (typically $G_i = 2^{\psi_i} - 1$ where $\psi_i$ is the relevance label), and $D$ is the rank discount function (typically $D(\tau(i)) = \log(1 + \tau(i))$). Here, $\tau(i)$ denotes the rank position of $y_i$ induced by the score $s$. 

This objective can optimize the DCG metric by prioritizing the placement of high-relevance items at the top of the list by applying a logarithmic discount to lower ranks. Consequently, any rank inversion that displaces a desirable response from its superior position results in a significant reduction of the overall score, which allows the $\Delta_{i,j}$ to impose a stringent penalty on misalignment that disrupts the intended preference order. 

\subsection{CRPO: Cross-lingual Ranking Preference Optimization}
\label{sec:crpo}
Building upon the LambdaLoss framework, CRPO aims to induce both cross-lingual and intra-lingual alignment by jointly optimizing target language responses alongside semantically equivalent preference knowledge from English. For each training instance, we construct a quadruple of responses consisting of: $y_w^t$, $y_l^t$, $y_w^e$, and $y_l^e$, which are chosen and rejected responses in the target language and English, respectively. To ensure a consistent preference manifold is maintained across languages, we assign hierarchical relevance labels~($\psi$) based on both response quality and language type among these four responses:
\begin{equation} 
\psi(y_w^t) > \psi(y_w^e) > \psi(y_l^t) > \psi(y_l^e)
\label{eq:crpo_hierarchy}
\end{equation}
Under this hierarchy, the model is trained to prioritize the preferred response in the target language, $y_w^t$, as the highest-ranked candidate. At the same time, CRPO provides comprehensive supervisory signals by aligning all pairs within the $\{y^t, y^e\}$ set for a given target language prompt $x$. The model is compelled to jointly optimize both intra-lingual preferences~($(y_w^t, y_l^t)$, $(y_w^e, y_l^e)$) and cross-lingual preference pairs, such as $(y_w^t, y_l^e)$ and $(y_w^e, y_l^t)$. 

Since the optimization of target language involves concurrent alignment with semantically equivalent English pairs, the model can leverage these English candidates as informative cues. Furthermore, prioritizing $y_w^t$ provides training signals for maintaining input-output language consistency. Thus, the gain function with $\psi$ in CRPO enables the model to inherit the logical consistency of English preferences while imposing a stringent quality constraint on responses within the target language. The gain functions for each response in CRPO are detailed in Appendix~\ref{appen:gain}.

Finally, we employ a negative log-likelihood~(NLL) loss term for $y_w^t$ in our loss to improve alignment performance during the instruction tuning process. The resulting loss function of CRPO is defined as:
\begin{equation}
\begin{aligned}
&\mathcal{L}_{\text{CRPO}}= (1-\alpha) \cdot \mathcal{L}_{\text{NLL}} + \alpha \cdot \mathcal{L}_{\text{lambda}}(\pi_\theta; \pi_{\text{ref}}, \beta) 
\end{aligned}
\end{equation}
where $\alpha$ is a hyperparameter that balances the ranking-based alignment and the language modeling objective. While conventional DPO is restricted to binary comparisons of responses within the target language, CRPO promotes the precise mapping of hierarchies defined by both language and quality onto a unified optimization plane.

\subsection{Weighting Schemes for CRPO}
\label{sec:weighting}
The flexibility of the LambdaLoss framework allows CRPO to incorporate various weighting functions for each optimizing different bounds of the ranking metric. We explore three schemes for $\Delta_{i,j}$ derived from LambdaLoss to investigate the effectiveness in cross-lingual alignment. 

\paragraph{LambdaRank} optimizes a coarse upper bound of the nDCG metric. It defines the weight based on the absolute difference between the inverse discounts of the two positions, which is presented in Equation~\ref{eq:lambda_weight}. This scheme provides a foundational gradient signal for ranking by considering the global positions of the responses in the list.

\paragraph{nDCG2} introduces a tighter bound for nDCG optimization. Unlike the global position-based weighting of LambdaRank, it focuses on the distance between the two items being compared:
\begin{equation}
\small
\begin{aligned}
&\Delta_{i,j}^{\text{nDCG2}} = \\
&|G_i - G_j| \cdot \left| \frac{1}{D(|\tau(i) - \tau(j)|)} - \frac{1}{D(|\tau(i) - \tau(j)| + 1)} \right|
\end{aligned}
\end{equation}
where $|\tau(i) - \tau(j)|$ represents the rank distance between the two responses. This scheme has been empirically shown to achieve superior performance by providing more localized and precise gradients for rank inversion.

\paragraph{nDCG2++} is a hybrid scheme that combines the strengths of both nDCG2 and LambdaRank to maximize ranking performance:
\begin{equation}
\Delta_{i,j}^{\text{nDCG2++}} = \mu \cdot \Delta_{i,j}^{\text{nDCG2}} + \Delta_{i,j}
\end{equation}
where $\mu$ is a hyperparameter that controls the contribution of the nDCG2 component. In our experiments, $\mu$ is set to 10 following the prior study~\cite{wang2018lambdaloss}.

While nDCG2++ is known to reach the strongest performance in traditional Information Retrieval tasks, we found that nDCG2 or LambdaRank alone can be sufficient for aligning LLM preference distributions across languages. To demonstrate the generality and theoretical robustness of the CRPO framework, we employ nDCG2 as our primary weighting scheme for the experiments.

\section{Experimental Setup}
\subsection{Models and Training}
\paragraph{Models}
We perform preference optimization with three different models, Llama-2-7B~\cite{touvron2023llama}, Llama-3-8B~\cite{dubey2024llama}, and Mistral-7B-v0.1~\cite{jiang2023mistral}. We employ five languages with varying resource availability: Chinese~(zh), Indonesian~(id), Korean~(ko), Swahili~(sw), and Bengali~(bn). They were chosen to provide a mix of high-, medium-, and low-resource languages, typological and script diversity, while satisfying the practical constraints of available evaluation datasets.

\paragraph{Dataset}
To construct the preference dataset for each language, we sample 3,000 instances from the UltraFeedback dataset~\cite{cui2024ultrafeedback} and translate them into the respective target languages using \texttt{gpt-5-chat}. This procedure yields fully parallel target language pairs $(x_t, y_w^t, y_l^t)$ that correspond directly to the original English preference pairs $(x_e, y_w^e, y_l^e)$. We randomly split the resulting dataset, including English pairs, into a 90\% training set and a 10\% test set.

\paragraph{Baseline}
We compare CRPO against two competitive baselines that aim to align cross-lingual preferences via monolithic strategy. We establish SFT+DPO as our strong baseline. In this setup, the model is optimized using response pairs in the language of the input. Since the training dataset encompasses both English and the target language, the model is trained on $(y_w^t, y_l^t)$ for $x^t$ and $(y_w^e, y_l^e)$ for $x^e$, in the same batch.

As another method, CLO~\cite{lee-etal-2025-cross} is a variant of DPO that incorporates binary comparison between English and target language response pairs within the same batch. To be specific, CLO explicitly steers the model toward generating desirable and language-consistent responses by designating the target language response as chosen and the English response as rejected. Furthermore, it mitigates inherent English bias and facilitates alignment by restricting the NLL loss exclusively to target language responses.

Note that all baselines and models use identical hyperparameters and the same amount of parallel datasets to ensure a fair comparison. Further details about the training can be found in Appendix~\ref{appen:hyper}.

\subsection{Evaluation}
\paragraph{AlpacaEval} 
We employ AlpacaEval~\cite{alpaca_eval} to evaluate models' conversational ability across a diverse set of queries. AlpacaEval comprises 805 questions from 5 datasets and evaluates model performance through qualitative comparisons of reference responses and candidate model outputs. To support evaluation across multiple languages, we adopt the multilingual version of AlpacaEval established in prior work~\cite{lee-etal-2025-cross} as our primary evaluation suite. We report the win rate~(WR) and length-controlled win rate~(LC) against the SFT model. The LC metric is specifically designed to be robust against model verbosity~\cite{dubois2024length}. Also, by including the instruction in the evaluation prompt, the model is penalized for responding in a language other than the target language. We use \texttt{gpt-5-chat} as a judge for assessment and also report multilingual Arena-Hard results in Appendix~\ref{app:arena}.

\paragraph{Knowledge \& Understanding}
We further evaluate the models' capacity for knowledge utilization and linguistic comprehension. We use MMMLU~(Multilingual Massive Multitask Language Understanding)~\cite{hendrycks2021measuringmassivemultitasklanguage} to verify whether the trained model can leverage inherent world knowledge in target language contexts. MMMLU is a human-translated multilingual extension of the original MMLU benchmark, encompassing 57 subjects across STEM, humanities, and social science \footnote{\url{https://huggingface.co/datasets/openai/MMMLU}}. In addition, we employ Belebele~\cite{bandarkar-etal-2024-belebele}, a benchmark for multilingual reading comprehension. In this task, we can evaluate literacy and contextual understanding, as the model should select the most appropriate answer from four choices based on the short passage and question. Following standard practice, we report accuracy by treating the option with the highest overall log-likelihood as the model's predicted answer~\cite{gao2021framework}.

\paragraph{In-depth Analysis}
To move beyond aggregate benchmark scores and gain deeper insights into the model's internal mechanics, we conduct an intrinsic analysis. We investigate the reward distributions and log probabilities of the preference pairs in our test set to verify whether each method enables balanced alignment that reinforces desirable generative behaviors, rather than relying solely on suppressing suboptimal outputs.

Furthermore, we measure the quality distribution of generated responses using an external reward model. The objective of this analysis is to empirically demonstrate that the internal optimization of CRPO translates into measurable improvements in output quality, ensuring that the model’s preference manifold is robustly aligned with objective judgments in the target language.

\begin{table*}[htb!]
\centering
\resizebox{\textwidth}{!}{
\begin{tabular}{ll cccccccc cc}
\toprule
\multirow{2}{*}{\textbf{Model}} & \multirow{2}{*}{\textbf{Method}} & \multicolumn{2}{c}{\textbf{Chinese}} & \multicolumn{2}{c}{\textbf{Indonesian}} & \multicolumn{2}{c}{\textbf{Korean}} & \multicolumn{2}{c}{\textbf{Swahili}} & \multicolumn{2}{c}{\textbf{Bengali}} \\
\cmidrule(lr){3-4} \cmidrule(lr){5-6} \cmidrule(lr){7-8} \cmidrule(lr){9-10} \cmidrule(lr){11-12}
& & LC~(\%) & WR~(\%) & LC~(\%) & WR~(\%) & LC~(\%) & WR~(\%) & LC~(\%) & WR~(\%) & LC~(\%) & WR~(\%) \\
\midrule
\multirow{3}{*}{Llama-2-7B} & SFT+DPO & 57.80 & 58.50 & 57.92 & 58.88 & 54.28 & 55.40 & 38.63 & 38.07 & 34.90 & 35.68 \\
& CLO & 56.41 & 56.08 & 53.84 & 53.91 & 51.15 & 50.24 & \textbf{44.49} & \textbf{43.97} & 33.15 & 30.93 \\
& CRPO & \textbf{64.18} & \textbf{64.34} & \textbf{63.83} & \textbf{64.47} & \textbf{63.02} & \textbf{63.97} & 44.07 & 43.72 & \textbf{37.52} & \textbf{36.77} \\
\midrule
\multirow{3}{*}{Llama-3-8B} & SFT+DPO & 60.24 & 60.37 & 59.24 & 60.37 & 64.97 & 65.96 & 46.23 & 46.95 & 34.77 & 36.89 \\
& CLO & 58.70 & 58.13 & 52.42 & 52.96 & 53.94 & 53.41 & 43.78 & 44.93 & 35.67 & 33.29 \\
& CRPO & \textbf{62.45} & \textbf{62.36} & \textbf{67.97} & \textbf{68.40} & \textbf{68.45} & \textbf{68.81} & \textbf{61.84} & \textbf{62.17} & \textbf{53.86} & \textbf{55.65} \\
\midrule
\multirow{3}{*}{Mistral-7B} & SFT+DPO & 55.65 & 56.14 & 55.98 & 58.50 & 51.49 & 52.85 & 27.95 & 27.33 & 24.63 & 24.59 \\
& CLO & 55.18 & 55.03 & 51.29 & 51.98 & 48.14 & 48.32 & 35.15 & 35.40 & 19.98 & 18.01 \\
& CRPO & \textbf{62.83} & \textbf{63.10} & \textbf{66.81} & \textbf{68.19} & \textbf{62.90} & \textbf{63.72} & \textbf{50.76} & \textbf{51.98} & \textbf{48.69} & \textbf{48.57} \\
\bottomrule
\end{tabular}}
\caption{AlpacaEval across five target languages. LC and WR denote length-controlled and raw win rate, respectively. The best results for each model are highlighted in \textbf{bold}.}
\label{tab:alpaca}
\end{table*}

\begin{table}[htb!]
\centering
\resizebox{\linewidth}{!}{
\begin{tabular}{ll ccc}
\toprule
\textbf{Model} & \textbf{Lang} & \textbf{SFT+DPO} & \textbf{CLO} & \textbf{CRPO} \\
\midrule
\multicolumn{5}{c}{\textbf{MMMLU}} \\
\midrule
\multirow{3}{*}{Llama-2-7B}
& id & 26.33 & 24.69 & \textbf{27.85} \\
& ko & 26.20 & 28.57 & \textbf{31.07} \\
& sw & 25.08 & \textbf{25.62} & 25.12 \\
\midrule
\multirow{3}{*}{Llama-3-8B}
& id & 41.69 & 36.41 & \textbf{45.94} \\
& ko & 39.58 & \textbf{40.19} & 39.84 \\
& sw & 36.02 & 27.48 & \textbf{37.03} \\
\midrule
\multirow{3}{*}{Mistral-7B}
& id & 37.36 & 31.38 & \textbf{37.47} \\
& ko & 30.39 & 29.27 & \textbf{35.01} \\
& sw & 30.57 & 28.32 & \textbf{31.78} \\
\midrule
\multicolumn{5}{c}{\textbf{Belebele}} \\
\midrule
\multirow{3}{*}{Llama-2-7B}
& id & 33.11 & 36.55 & \textbf{36.66} \\
& ko & 33.00 & 35.66 & \textbf{36.33} \\
& sw & 29.22 & 28.22 & \textbf{29.22} \\
\midrule
\multirow{3}{*}{Llama-3-8B}
& id & 64.22 & 35.11 & \textbf{65.88} \\
& ko & 57.55 & 59.44 & \textbf{68.66} \\
& sw & 47.55 & 28.66 & \textbf{49.22} \\
\midrule
\multirow{3}{*}{Mistral-7B}
& id & \textbf{48.33} & 30.44 & 47.44 \\
& ko & 49.66 & 35.77 & \textbf{51.77} \\
& sw & 33.00 & 33.88 & \textbf{34.77} \\
\bottomrule
\end{tabular}
}
\caption{MMMLU and Belebele evaluation results. We report the performance of zero-shot in MMMLU and one-shot in Belebele.}
\label{tab:mmlu_bele}
\end{table}

\section{Results and Analysis}
In this section, we present the main results of our experiments, highlighting CRPO's superior performance across various benchmarks in Section~\ref{sec:main}. Then in Section~\ref{sec:analysis}, we provide an in-depth understanding of the internal reward dynamics and generative behaviors that distinguish CRPO from other alignment methods. In addition, we employ an external reward model to objectively quantify improvements in absolute generation quality in Section~\ref{sec:external}. Finally, we conduct a study on various ranking schemes to investigate how different configurations of the hierarchical preference structure influence the effectiveness of cross-lingual alignment in Section~\ref{sec:ranking}.

\subsection{Main Results}
\label{sec:main}
\paragraph{Overall Performance}
As shown in Table~\ref{tab:alpaca}, all preference optimization methods provide improvements over the SFT model in most cases, with CRPO achieving the best results across languages. SFT+DPO, a standard approach for model alignment, also serves as a robust baseline in other languages. CLO exhibits comparable or slightly below performance to SFT+DPO, demonstrating the utility of a straightforward objective that suppresses the probability of English responses. Furthermore, we include the English performance of each case in Appendix~\ref{appen:tax} in terms of alignment tax.

\paragraph{Robustness in Low-Resource Languages}
While most methods outperform SFT in high- and mid-resource languages, a different tendency emerges in low-resource settings. Unlike CRPO, other methods exhibit notable performance degradation in low-resource scenarios. This trend is particularly pronounced for Llama-3-8B, which suffers from a collapse compared to the SFT, as evidenced by notably low WR in Bengali.

\begin{figure*}[htb!]
\centering
\subcaptionbox{Reward Difference}{\includegraphics[width=\textwidth]{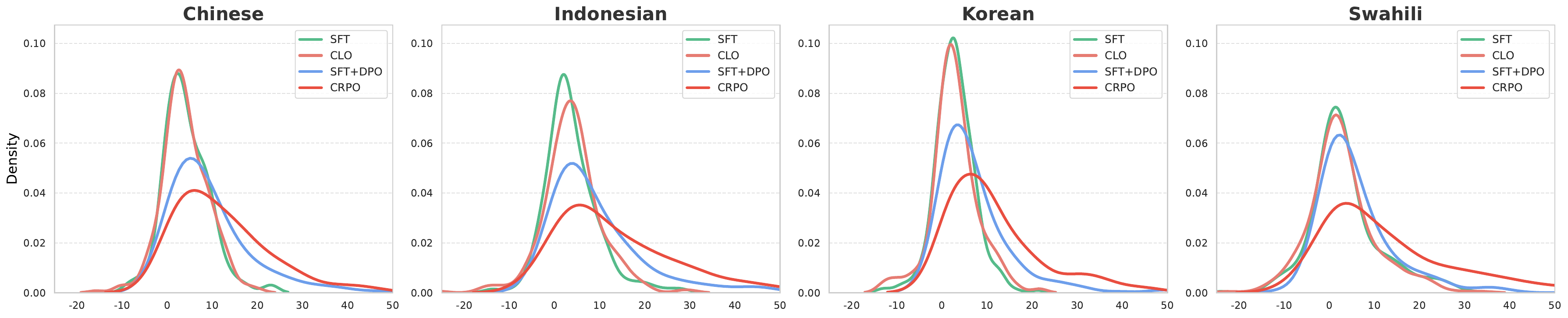}}
\vfill
\subcaptionbox{Log Probability Distribution}{\includegraphics[width=\textwidth]{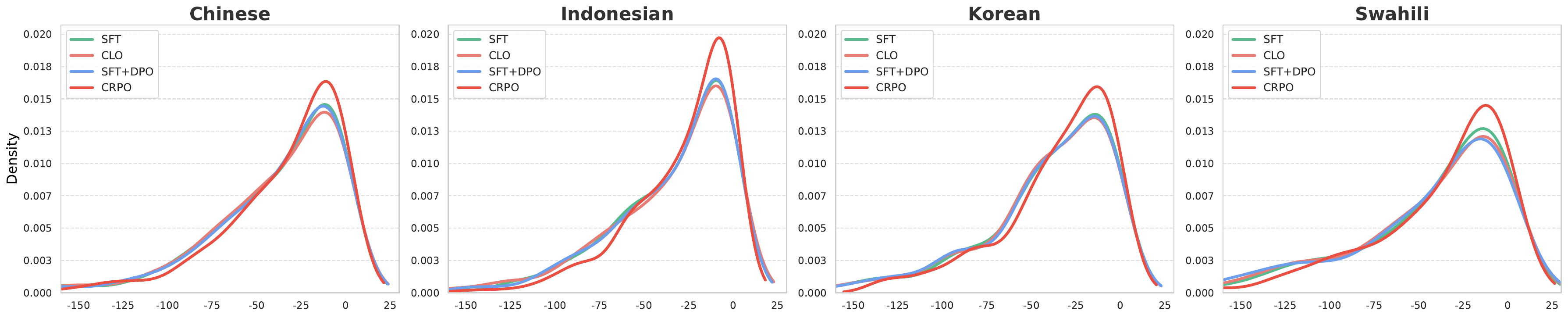}}
\caption{(a) Reward difference distribution and (b) Log-likelihood distribution on chosen responses~($y_w$) of Llama-3-8B trained with~SFT (blue), CLO~(orange), SFT+DPO~(blue) and CRPO~(red) across languages.}
\label{fig:dist}
\end{figure*}

This stems from optimization instability and excessive parameter volatility during the preference alignment when the model's linguistic grounding in the target language is insufficient~\cite{kadyrbek2025development,lee-etal-2025-cross}. From this perspective, SFT+DPO, relying on isolated preference pairs in the target language, fails to establish a stable preference manifold in low-resource scenarios. Similarly, the performance drop in CLO shows that simply enforcing language-consistent responses through binary cross-lingual hierarchies is insufficient for robust alignment.

While CRPO also encounters adaptation challenges due to the inherent limitations of models, it consistently surpasses baselines and demonstrates remarkable robustness in most cases. Notably, CRPO achieves a significant performance gain in Swahili with Llama-3, scoring a WR of 62.17. This gain can be attributed to the use of relatively well-aligned English pairs as a logical anchor for aligning the target language. By realigning language quality and hierarchy, CRPO exhibits the most robust resistance to this alignment collapse and contributes to overall alignment stabilization.

\begin{table}[t]
\centering
\resizebox{\linewidth}{!}{
\begin{tabular}{ll cccc}
\toprule
\textbf{Model} & \textbf{Method} & \textbf{zh} & \textbf{id} & \textbf{ko} & \textbf{sw} \\
\midrule
\multirow{4}{*}{Llama-2-7B}
& SFT    & -1.749 & -1.797 & -2.870 & -2.901 \\
& SFT+DPO & \underline{-1.227} & \textbf{-0.904} & \underline{-2.121} & -2.958 \\
& CLO    & -1.539 & -1.200 & -2.855 & \underline{-2.457} \\
& CRPO   & \textbf{-0.607} & \underline{-0.936} & \textbf{-1.601} & \textbf{-2.175} \\
\midrule
\multirow{4}{*}{Llama-3-8B}
& SFT    & 0.344 & 0.208 & -0.904 & -0.487 \\
& SFT+DPO & \underline{1.506} & \underline{1.585} & \underline{0.869} & \underline{-0.212} \\
& CLO    & 0.613 & 1.034 & -0.119 & -0.531 \\
& CRPO   & \textbf{1.738} & \textbf{1.883} & \textbf{0.911} & \textbf{1.122} \\
\midrule
\multirow{4}{*}{Mistral-7B}
& SFT    & -0.706 & -0.808 & -1.800 & \underline{-0.667} \\
& SFT+DPO & \underline{0.255} & \underline{-0.037} & \underline{-0.719} & -1.519 \\
& CLO    & -0.096 & -0.665 & -1.237 & -0.981 \\
& CRPO   & \textbf{0.508} & \textbf{0.805} & \textbf{-0.344} & \textbf{-0.040} \\
\bottomrule
\end{tabular}
}
\caption{Average of reward scores from Skywork-Reward-V2-Qwen3-8B on generation outputs. The best results achieved across the methods are highlighted in bold, while the second-best results are underlined.}
\label{tab:reward_score}
\end{table}

\paragraph{Knowledge Utilization}
The validity of CRPO extends beyond conversational improvements, as evidenced in the knowledge utilization and linguistic comprehension areas. As shown in Table~\ref{tab:mmlu_bele}, CRPO achieves competitive performance across both benchmarks. Notably, the MMMLU score in Indonesian for Llama-3 exceeds the strongest baseline by over 4 points, while the Korean Belebele score reaches 68.66. These results demonstrate that the ranking structure in CRPO successfully anchors the model’s inherent knowledge system within the target language space, thereby achieving better alignment for complex reasoning tasks.

\subsection{Reward and Probability Shifts}
\label{sec:analysis}
To investigate the impact of CRPO on the model’s preference manifold, we examine the internal dynamics of reward differences and generative log-probabilities. As illustrated in Figure~\ref{fig:dist}, the magnitude of distribution shifts differs across methods regarding the SFT model as a default.

In Figure~\ref{fig:dist}~(a), we find that regardless of the language, CRPO consistently achieves a large reward difference and shifts the distribution in a positive direction compared to the SFT. Likewise, the SFT+DPO approach exhibits a similar trend, though the magnitude of the shift is relatively smaller. In contrast, CLO yields almost identical reward difference to SFT, or in the case of Indonesian, even shows signs of degradation.

\begin{figure*}[htb!]
\centering
\includegraphics[width=0.85\textwidth]{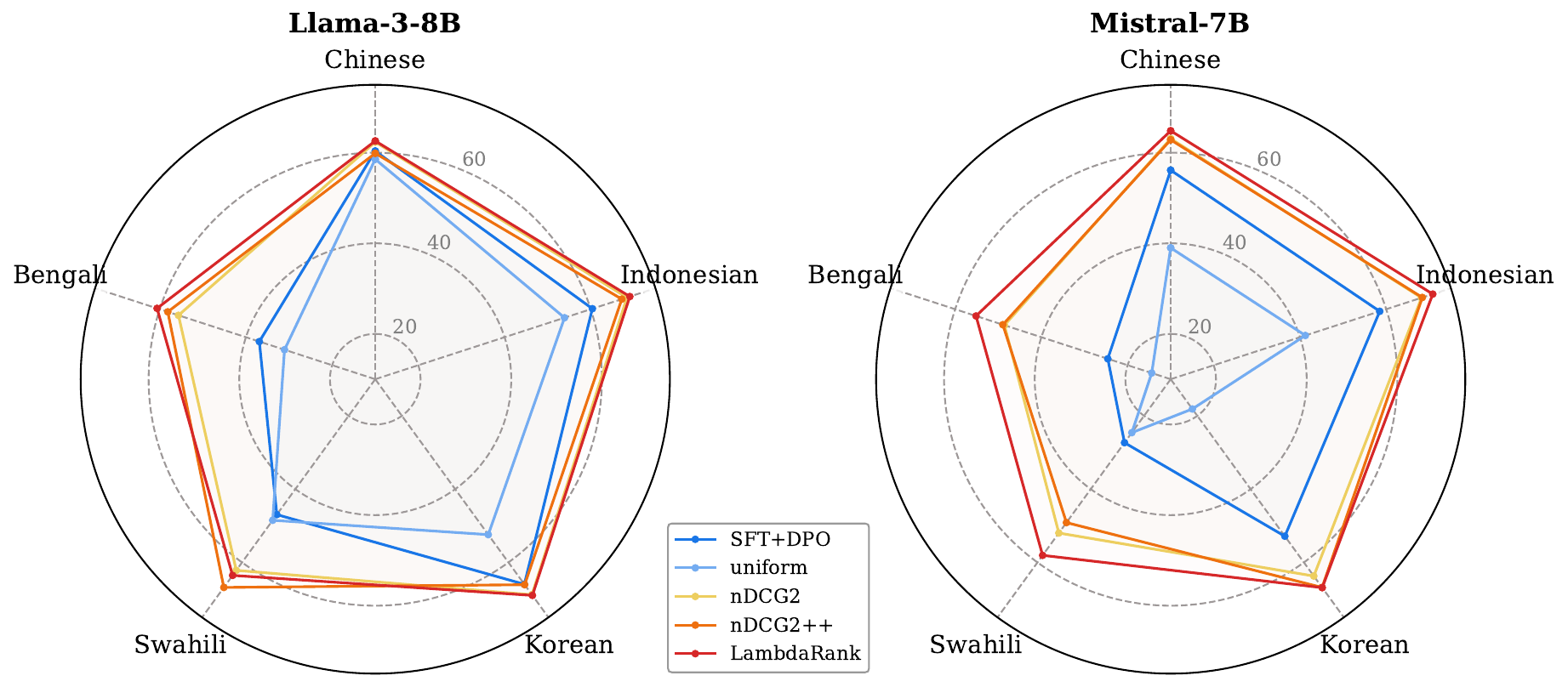}
\caption{Win rate comparison of different weighting schemes within the CRPO framework.}
\label{fig:radar}
\end{figure*}

One of the primary risks in preference optimization is the improper focus on maximizing the reward margin, which often results in suppressing the probability of rejected responses rather than increasing the probability of chosen responses. This imbalance can undermine the language model's inherent stability and fluency by failing to provide an explicit guide for target responses, potentially leading to degeneration~\cite{meng2024simpo,hong2024orpo}. As shown in the comparison of Figure~\ref{fig:dist}~(a) and (b), while other methods increase the reward margin, their log-likelihood distributions for chosen responses remain largely congruent with the SFT. This indicates that other approaches primarily fail to incentivize the generation of favored responses.

On the other hand, CRPO assigns higher values to both the log-likelihood of the chosen response and the reward difference across all languages than in other cases. This demonstrates that CRPO achieves better alignment in the target language by increasing the likelihood of the favored response. Further analysis of the reward distribution is provided in Appendix~\ref{appen:dist}.

\subsection{External Quality Assessment}
\label{sec:external}
While win rates provide a measure of relative preference, they may not fully capture the absolute quality shift in the model's generative distribution. To provide a more objective assessment of response quality, we use Skywork-Reward-V2-Qwen3-8B~\cite{liu2025skywork}, an external reward model known to support multilingual environments, to score the outputs of each method on the test set. Table~\ref{tab:reward_score} summarizes the average reward scores across diverse models and languages.

The results demonstrate that CRPO achieves superior reward magnitudes across nearly all configurations. A notable score escalation is observed in the Mistral-7B; in Indonesian~(id), while the SFT+DPO reaches a score of -0.037, CRPO elevates the quality to a considerable positive margin of 0.805, outperforming SFT+DPO by a wide gap. This also substantiates our claim that anchoring target language alignment to established English preference structures can guide the model toward higher-quality regions of the generation manifold.

\subsection{Study on Ranking Scheme}
\label{sec:ranking}
We investigate how the choice of weighting scheme in Equation~\ref{eq:lambda_loss} influences cross‑lingual preference alignment. Figure~\ref{fig:radar} shows AlpacaEval results obtained by diverse weighting schemes within the same CRPO framework.

As shown in Figure~\ref{fig:radar}, all three schemes consistently outperform both SFT+DPO and the uniform scheme, confirming that hierarchical ranking signals provide more informative supervision than pairwise comparisons. We can attribute the success across diverse configurations to CRPO's hierarchical structural design. As evidence for this, assigning a uniform weight to all response pairs results in a notable performance drop. The degradation mainly arises from the lack of alignment between the languages. Removing hierarchical cues triggers an English bias, leading to a failure in linguistic consistency with the given prompt.

Moreover, LambdaRank and nDCG2 achieve strong performance across most settings. This suggests that global position‑based signals and local distance‑based signals can complement each other depending on the model and language. Given that our experiments are conducted under the nDCG2 scheme, the additional gains observed with alternative schemes further indicate that the potential performance ceiling of CRPO may be even higher.

\section{Conclusion}
In this paper, we propose CRPO, a new alignment framework designed to establish robust preference manifolds both within and across languages. By reformulating the target language adaptation and alignment task as a hierarchical ranking problem, CRPO benefits from the transfer of preference knowledge along the language chain. Our empirical results demonstrate that CRPO not only achieves consistent superiority over standard pairwise approaches but also enhances the generative likelihood of preferred responses, facilitating the generation of high-fidelity, precise responses in the target language. We believe that our approach shifts the paradigm of cross-lingual alignment from isolated binary comparisons to an integrated cross-lingual ranking perspective. Moving forward, we will investigate a dynamic gain adaptation strategy based on linguistic similarities and training stages. By mitigating the optimization complexity inherent in expansive ranking spaces, we expect to ensure robust cross-lingual alignment across an even wider array of language pairs.

\section*{Limitations}
Although we use the same amount of the dataset during the training, our proposed framework employs a hierarchical ranking structure that processes four response candidates in a single step. This inherently demands more computational resources during training. However, this overhead can be considered an unavoidable cost inherent to methods that leverage item-wise ranking, not only limited to the preference optimization field. Given the limitations of standard binary comparisons in optimizing both quality and consistency simultaneously, the performance gains achieved by CRPO demonstrate that this additional cost is a worthwhile trade-off for effective cross-lingual alignment.

Furthermore, our main experiments include evaluation results from the multilingual version of the AlpacaEval dataset, translated using gpt-4o in prior work~\cite{lee-etal-2025-cross}, and the MMMLU dataset~\cite{hendrycks2021measuringmassivemultitasklanguage}, which was translated into the respective languages by professional human translators. Additionally, we employed the Belebele dataset~\cite{bandarkar-etal-2024-belebele}, which was constructed and reviewed by human experts during translation. While these datasets are suitable for measuring general model performance, they may not fully capture the model's ability to respond appropriately to queries involving language-specific characteristics or cultural contexts. Although this limitation may prevent a comprehensive evaluation of the model's grasp of cultural nuances and localized content, we made every effort to utilize the highest-quality benchmarks available.

\section*{Ethics Statement}
In this research, we utilized only publicly available resources. All training and evaluation data were obtained from publicly authorized open-access repositories. Nevertheless, we recognize the potential for residual biases or harmful concepts in the English source data to be propagated across languages. To mitigate this, we carefully adhered to the copyright guidelines and terms of use for all original works, datasets, and language resources, including translated materials. We confirm that the collection, processing, and application of these resources raise no distinct ethical concerns.

\section*{Acknowledgments}
This research was supported by Basic Science Research Program through the National Research Foundation of Korea~(NRF) funded by the Ministry of Education~(NRF-2021R1A6A1A03045425). This work was supported by Institute for Information \& communications Technology Promotion~(IITP) grant funded by the Korea government~(MSIT)~(RS-2024-00398115, Research on the reliability and coherence of outcomes produced by Generative AI). This research was supported by Culture, Sports and Tourism R\&D Program through the Korea Creative Content Agency grant funded by the Ministry of Culture, Sports and Tourism in 2026 (Project Name: Development of an AI Agent Integrating Korean Language Knowledge for Personalized Language Consultation Services, Project Number: RS-2026-25506607, Contribution Rate: 25\%). This work was supported by Institute of Information \& communications Technology Planning \& Evaluation~(IITP) under the Leading Generative AI Human Resources Development~(IITP-2026-RS-2026-25545512) grant funded by the Korea government~(MSIT).

\bibliography{custom}

\appendix

\section{Gain Formulation in Cross-lingual Ranking}
\label{appen:gain}

\paragraph{Preference Hierarchy}
In our pilot study, we observed that language adaptation is more readily achieved than the fine-grained differentiation of response quality within the same language, which can easily overshadow quality optimization during training. To reflect this, we established the following preference hierarchy for our ranking-based objective: $\psi(y_w^t) > \psi(y_w^e) > \psi(y_l^t) > \psi(y_l^e)$. A critical design choice in this formulation is the relation $\psi(y_w^e) > \psi(y_l^t)$, which prioritizes a high-quality response in English over a lower-quality response in the target language. Furthermore, this provides a learning signal that encourages the model to generate desirable responses regardless of the language. Since gain adjustments across items in a ranking structure are interdependent rather than orthogonal, penalizing the English winner below the target language loser would otherwise heavily bias the learning signal toward language matching.


\begin{table*}[htb!]
\centering
\resizebox{\linewidth}{!}{
\begin{tabular}{lcccccccccc}
\toprule
\multirow{2}{*}{\textbf{Gain values~($G$)}} & \multicolumn{2}{c}{\textbf{Chinese}} & \multicolumn{2}{c}{\textbf{Korean}} & \multicolumn{2}{c}{\textbf{Indonesian}} & \multicolumn{2}{c}{\textbf{Swahili}} & \multicolumn{2}{c}{\textbf{Bengali}} \\
\cmidrule(lr){2-3} \cmidrule(lr){4-5} \cmidrule(lr){6-7} \cmidrule(lr){8-9} \cmidrule(lr){10-11}
& WC~(\%) & LC~(\%) & WC~(\%) & LC~(\%) & WC~(\%) & LC~(\%) & WC~(\%) & LC~(\%) & WC~(\%) & LC~(\%) \\
\midrule
7, 3, 1, 0 & 60.50 & 60.75 & 66.46 & 66.04 & 67.20 & 66.86 & 66.46 & 66.35 & 54.16 & 52.54 \\
9, 5, 7, 4 & 60.00 & 60.39 & 66.21 & 66.11 & 65.22 & 65.52 & \textbf{67.14} & \textbf{67.43} & 53.98 & 53.79 \\
9, 7, 5, 4 & \textbf{62.36} & \textbf{62.46} & \textbf{68.82} & \textbf{68.46} & \textbf{68.41} & \textbf{67.98} & 62.17 & 61.85 & \textbf{55.65} & \textbf{53.87} \\
\bottomrule
\end{tabular}
}
\caption{WR and LC according to different gain values.}
\label{tab:gain_comp}
\end{table*}

\begin{table*}[htb!]
\centering
\resizebox{\textwidth}{!}{
\begin{tabular}{llcccccccccc}
\toprule
\multirow{2}{*}{\textbf{Model}} & \multirow{2}{*}{\textbf{Method}} & \multicolumn{2}{c}{\textbf{Chinese}} & \multicolumn{2}{c}{\textbf{Indonesian}} & \multicolumn{2}{c}{\textbf{Korean}} & \multicolumn{2}{c}{\textbf{Swahili}} & \multicolumn{2}{c}{\textbf{Bengali}} \\
\cmidrule(lr){3-4} \cmidrule(lr){5-6} \cmidrule(lr){7-8} \cmidrule(lr){9-10} \cmidrule(lr){11-12}
& & LC~(\%) & WR~(\%) & LC~(\%) & WR~(\%) & LC~(\%) & WR~(\%) & LC~(\%) & WR~(\%) & LC~(\%) & WR~(\%) \\
\midrule
\multirow{3}{*}{Llama-2-7B} & SFT+DPO & 55.06 & 55.71 & 55.89 & 56.52 & 57.79 & 58.26 & 56.04 & 56.46 & 51.46 & 52.42 \\
& CLO & 57.46 & 57.27 & 53.49 & 52.24 & 55.30 & 54.91 & 53.41 & 51.93 & 53.22 & 52.61 \\
& CRPO & \textbf{62.33} & \textbf{62.36} & \textbf{60.41} & \textbf{60.62} & \textbf{60.46} & \textbf{60.87} & \textbf{60.43} & \textbf{60.00} & \textbf{61.60} & \textbf{61.99} \\
\midrule
\multirow{3}{*}{Llama-3-8B} & SFT+DPO & \textbf{57.71} & \textbf{58.45} & 60.31 & 61.30 & 56.23 & 56.83 & 58.06 & 58.01 & 56.98 & \textbf{57.76} \\
& CLO & 52.83 & 52.48 & 56.89 & 56.21 & 52.66 & 50.99 & 54.46 & 52.61 & 56.27 & 54.78 \\
& CRPO & 56.73 & 57.20 & \textbf{64.24} & \textbf{64.35} & \textbf{59.19} & \textbf{59.07} & \textbf{60.53} & \textbf{60.06} & \textbf{57.43} & 57.64 \\
\midrule
\multirow{3}{*}{Mistral-7B} & SFT+DPO & \textbf{61.30} & \textbf{65.59} & 58.44 & 61.06 & \textbf{63.31} & \textbf{69.69} & 51.73 & 54.88 & 47.65 & 55.47 \\
& CLO & 51.84 & 54.47 & 52.05 & 51.61 & 58.19 & 61.80 & 50.37 & 49.81 & 47.97 & 45.83 \\
& CRPO & 50.60 & 57.08 & \textbf{59.96} & \textbf{61.86} & 57.42 & 64.35 & \textbf{56.21} & \textbf{57.76} & \textbf{53.49} & \textbf{55.86} \\
\bottomrule
\end{tabular}}
\caption{English performance in AlpacaEval benchmark. The best results for each model are highlighted in \textbf{bold}.}
\label{tab:alpaca_en}
\end{table*}

\paragraph{Balance between Two Dimensions}
This philosophy also led to our choice of gain values. In standard Information Retrieval, which adopts the LambdaLoss framework, the gain function typically follows an exponential discount: $G_i = 2^{\psi_i} - 1$~\cite{wang2018lambdaloss}. However, applying this to our dual-dimensional objective can introduce an inductive bias.

Given our hierarchy, the standard formula yields $G_w^t = 7$, $G_w^e = 3$, $G_l^t = 1$, and $G_l^e = 0$. Although ranking optimization inherently assigns disproportionate weights between items, this exponential spacing exacerbates biased local gradients.

To resolve this and provide a more balanced optimization manifold, we manually set the gains to $G_w^t = 9$, $G_w^e = 7$, $G_l^t = 5$, and $G_l^e = 4$. In this setup, the primary intra-language quality gaps ($G_w^t - G_l^t = 4$ for target, $G_w^e - G_l^e = 3$ for English) remain strictly greater than the primary language gap ($G_w^t - G_w^e = 2$). This constraint explicitly ensures that the LambdaLoss framework jointly optimizes both cross-lingual alignment and quality, preventing the simple language-matching signal from overwhelming the fine-grained quality optimization.

\paragraph{Empirical Validation}
To validate our gain formulation, we compare the performance of Llama-3-8B across different gain configurations, as presented in Table \ref{tab:gain_comp}. The theoretical range of possible gains is infinite, but as it is computationally prohibitive to explore the entire space, our experiments are restricted to a set of representative fixed values.

The standard exponential setting ($7, 3, 1, 0$) generally yields suboptimal win rates, suggesting that disproportionate local gradients hinder effective intra-lingual alignment. Furthermore, we examine a configuration ($9, 5, 7, 4$) that explicitly violates our preference hierarchy by penalizing the English winner ($G=5$) below the target language loser ($G=7$). This configuration also yields slightly lower performance than our hierarchy across most languages, including Chinese, Korean, Indonesian, and Bengali, demonstrating that heavily biasing the signal toward language matching compromises overall response quality. Interestingly, a different tendency appears in Swahili. We attribute this to the low-resource nature of Swahili, where an aggressive language-matching penalty might provide a temporary empirical advantage in enforcing the target script. Nevertheless, our proposed hierarchy configuration demonstrates greater robustness across diverse linguistic environments, justifying our balanced hierarchical design.

\section{Implementation Details}
\label{appen:hyper}
\paragraph{Training} We observe that hyperparameter tuning is crucial for achieving optimal performance of preference optimization methods. To ensure a
fair comparison, we conduct thorough hyperparameter tuning across all methods in our experiments.

\begin{figure*}[htb!]
\centering
\subcaptionbox{Reward distribution of $y_w$}{\includegraphics[width=\textwidth]{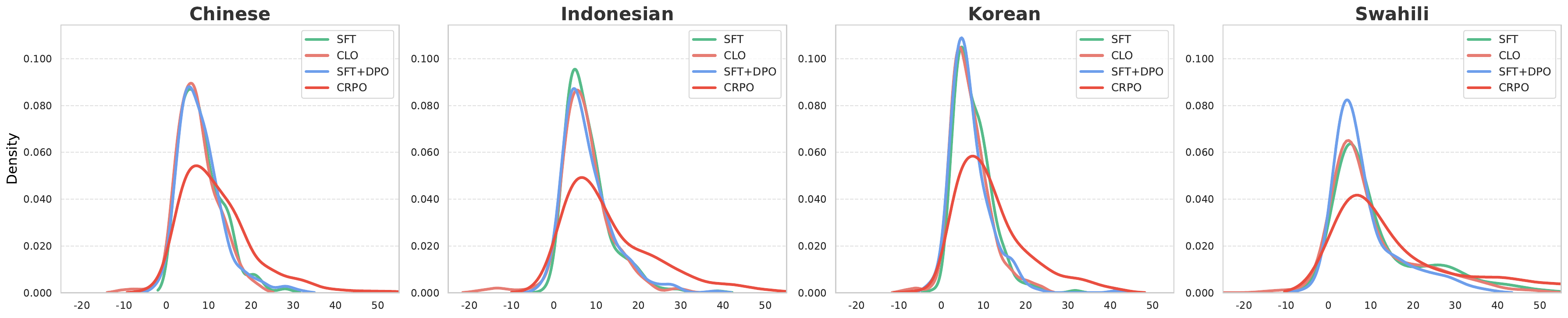}}
\vfill
\subcaptionbox{Reward distribution of $y_l$}{\includegraphics[width=\textwidth]{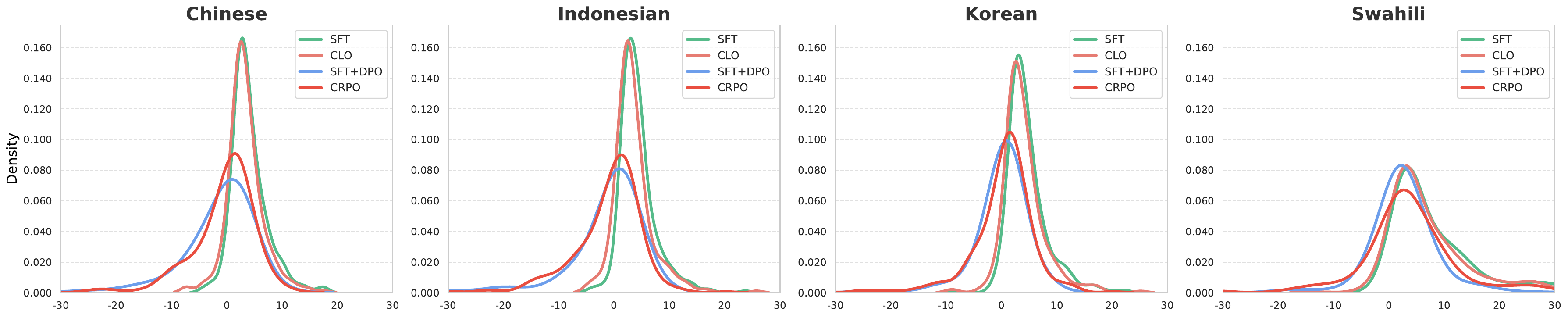}}
\caption{Reward distribution of chosen response~(a) and rejected response~(b) of Llama-3-8B trained with~SFT (blue), CLO~(orange), SFT+DPO~(blue) and CRPO~(red) across languages.}
\label{fig:reward_dist}
\end{figure*}

We conduct preliminary experiments to search for learning rate in $[5e-7, 8e-6, 1e-5]$ and training epochs in $[1, 2]$. We find that an effective learning rate of 8e-6 and training for two epochs consistently yield stable results across all models and languages. Therefore, we fix these values for all preference optimization experiments and report results using the final checkpoint. Additionally, we set the maximum sequence length to 3072 and apply a linear learning rate schedule with a 10\% warmup ratio. To balance the training objective, we use a weighting factor of $\alpha = 0.2$ for CRPO and SFT+DPO, and 0.5 for CLO following the recommended value in the original paper. We set $\beta$ to 0.1, the default value in DPO. All experiments are conducted under four NVIDIA A100 GPUs.

\paragraph{Evaluation} For generative evaluation, we maintain the same maximum sequence length as used during the training phase to ensure consistency. In the generation process, we use temperature as the primary hyperparameter in the decoding step and set it to 0.8 to balance the trade-off between coherence and diversity in the outputs.

\section{Arena-Hard}
\label{app:arena}
To further substantiate our evaluation, we report the multilingual Arena-Hard~(m-ArenaHard)\footnote{\url{https://huggingface.co/datasets/CohereLabs/m-ArenaHard}}~\cite{dang2024ayaexpansecombiningresearch} evaluation results in Table~\ref{tab:arena}. The m-ArenaHard benchmark was created by translating the prompts from the original Arena-Hard-v0.1~\cite{li2024crowdsourced} test dataset to 22 languages. We conduct supplementary experiments on the three languages that overlap with those covered in our study. Since several queries in the dataset exceed the relatively short context length of Llama-2 (4096 tokens), Llama-2 is excluded from the evaluation. Similar to AlpacaEval, we report the win rate against the SFT model.

\begin{table}[t]
\centering
\resizebox{\linewidth}{!}{
\begin{tabular}{ll ccc}
\toprule
\textbf{Model} & \textbf{Method} & \textbf{Chinese} & \textbf{Indonesian} & \textbf{Korean} \\
\midrule
\multirow{3}{*}{Llama-3-8B}
& SFT+DPO & 61.56 & 61.34 & 64.30 \\
& CLO     & 54.49 & 54.53 & 54.95 \\
& CRPO    & \textbf{62.15} & \textbf{61.87} & \textbf{68.81} \\
\midrule
\multirow{3}{*}{Mistral-7B}
& SFT+DPO & 54.62 & 54.34 & 55.39 \\
& CLO     & 55.49 & 53.34 & 53.02 \\
& CRPO    & \textbf{58.22} & \textbf{63.85} & \textbf{62.63} \\
\bottomrule
\end{tabular}
}
\caption{Win rate in Arena-Hard benchmark.}
\label{tab:arena}
\end{table}

\section{Impact on Alignment Tax}
\label{appen:tax}
In our main experiments, we focused on evaluating the target languages to demonstrate the effectiveness of cross-lingual alignment. However, since our approach leverages the model's inherent English knowledge as an anchor to induce transfer to the target language, it is imperative to investigate how the alignment training on the target language affects the model's English proficiency. To this end, we conduct an additional English evaluation on AlpacaEval to measure the alignment tax.

Table~\ref{tab:alpaca_en} presents the English performance of the models trained under each target language setting. Note that all methods also include English in their training data composition. As shown in the results, CRPO achieves higher English performance compared to the baselines in most cases, indicating that it successfully mitigates the alignment tax. To be specific, for the Llama-2-7B and Llama-3-8B models, CRPO achieves the highest WR across almost all target language configurations. This suggests that CRPO preserves the robustness of the model's English capabilities while effectively performing cross-lingual alignment.

\section{Reward Distribution Analysis}
\label{appen:dist}
In addition to Figure~\ref{fig:dist}, we also provide an analysis of the reward distribution of responses in Figure~\ref{fig:reward_dist}. As shown in Figure~\ref{fig:reward_dist}~(a), CRPO consistently yields higher rewards for favored responses compared to the other methods, whereas the reward distributions for the other methods do not exhibit a significant shift relative to SFT.

In Figure~\ref{fig:reward_dist}~(b), SFT+DPO and CRPO demonstrate comparable distributional shifts, which is the nature of DPO training that generally increases the reward of both chosen and rejected responses. Taken together, these findings indicate that the major reward difference achieved by CRPO, as observed in Figure~\ref{fig:dist}, is primarily driven by elevating the rewards of desirable responses rather than solely relying on penalizing those of unfavored ones. This further substantiates our claim in Section~\ref{sec:analysis} that the advantage of CRPO stems from promoting high-quality responses rather than solely from suppressing the probability of disfavored responses.

\section{Comparison with Another Ranked Optimization}
\label{sec:comparison_ranked_policy}
To investigate whether the effectiveness of CRPO can be attributed simply to the use of ranked preference structure, we further compare CRPO with PRO~\cite{song2024preference}, which constructs preferences sequentially from a ranked set of responses:
\begin{equation}
\mathcal{L}_{\mathrm{PRO}}
=
-\log
\prod_{k=1}^{n-1}
\frac{
\exp\left(r_{\pi}(x,y_k)\right)
}{
\sum_{i=k}^{n}\exp\left(r_{\pi}(x,y_i)\right)
},
\end{equation}
which is further combined with an NLL objective. Table~\ref{tab:pro_comparison} reports the AlpacaEval results across three target languages and two backbone models.

\begin{table}[t]
\centering
\small
\setlength{\tabcolsep}{3.2pt}
\begin{tabular}{lllcccc}
\toprule
\multirow{2}{*}{\textbf{Lang}} & \multirow{2}{*}{\textbf{Model}} & \multirow{2}{*}{\textbf{Method}}
& \multicolumn{2}{c}{\textbf{Target}} & \multicolumn{2}{c}{\textbf{English}} \\
\cmidrule(lr){4-5} \cmidrule(lr){6-7}
& & & WR & LC & WR & LC \\
\midrule

\multirow{4}{*}{zh}
& \multirow{2}{*}{Llama-3-8B}
& PRO  & 24.97 & 24.24 & 37.27 & 36.85 \\
& & CRPO & \textbf{62.36} & \textbf{62.45} & \textbf{57.20} & \textbf{56.73} \\
\cmidrule(lr){2-7}
& \multirow{2}{*}{Mistral-7B}
& PRO  & 13.42 & 13.75 & 36.15 & 33.67 \\
& & CRPO & \textbf{63.10} & \textbf{62.83} & \textbf{57.08} & \textbf{50.60} \\

\midrule

\multirow{4}{*}{id}
& \multirow{2}{*}{Llama-3-8B}
& PRO  & 23.11 & 24.79 & 44.47 & 43.60 \\
& & CRPO & \textbf{68.40} & \textbf{67.97} & \textbf{64.35} & \textbf{64.24} \\
\cmidrule(lr){2-7}
& \multirow{2}{*}{Mistral-7B}
& PRO  & 13.29 & 12.42 & 29.81 & 30.52 \\
& & CRPO & \textbf{68.19} & \textbf{66.81} & \textbf{61.86} & \textbf{59.96} \\

\midrule

\multirow{4}{*}{sw}
& \multirow{2}{*}{Llama-3-8B}
& PRO  & 11.43 & 11.78 & 37.64 & 36.17 \\
& & CRPO & \textbf{62.17} & \textbf{61.84} & \textbf{60.06} & \textbf{60.53} \\
\cmidrule(lr){2-7}
& \multirow{2}{*}{Mistral-7B}
& PRO  & 6.58 & 5.11 & 10.62 & 12.15 \\
& & CRPO & \textbf{51.98} & \textbf{50.76} & \textbf{57.76} & \textbf{56.21} \\

\bottomrule
\end{tabular}
\caption{AlpacaEval results of PRO and CRPO under the same training budget.}
\label{tab:pro_comparison}
\end{table}

Despite also performing ranking-based preference optimization, PRO substantially underperforms CRPO. An inspection of generations from the PRO-trained models shows that they generally retain the ability to respond in the language of the input. This suggests that PRO can achieve the primary objective of \emph{language adaptation}, but does not translate this adaptation into a comparable improvement in response quality. In contrast, CRPO improves both target language and English AlpacaEval performance by a large margin.

We attribute this difference to how the ranked preference structure is exploited during optimization. PRO primarily constructs pairwise comparisons between a higher-ranked response and lower-ranked alternatives. CRPO instead exploits the relative positions of responses throughout the ranked list and incorporates ranking-aware objectives to place greater emphasis on preference violations that affect the top of the ranking. Consequently, CRPO provides a more fine-grained learning signal for distinguishing responses with different quality levels rather than merely separating preferred from less-preferred responses.

This distinction is particularly important in our setting because language adaptation itself can already be encouraged by the NLL objective: the model can learn to generate in the target language without necessarily improving the quality of the generated content. The ranking-aware objective of CRPO complements this signal by explicitly preserving and promoting high-quality responses according to their positions in the ranked list. The considerable gap between PRO and CRPO therefore indicates that the gains of CRPO do not arise merely from introducing ranked preference optimization. Rather, \emph{how} the ranked preference information is incorporated into the optimization objective is critical, further supporting the design of CRPO.



\end{document}